\documentclass[conference]{IEEEtran}
\IEEEoverridecommandlockouts
\usepackage{cite}
\usepackage{amsmath,amssymb,amsfonts}
\usepackage[ruled,vlined]{algorithm2e}
\usepackage{graphicx}
\usepackage{pgfplots}
\pgfplotsset{width=6cm,compat=1.9}
\usepackage{pgfplotstable}
\usepackage[caption=true]{subfig}
\usepgfplotslibrary{fillbetween}
\usepackage{textcomp}
\usepackage{xcolor}
\usepackage{comment}
\usepackage{hyperref}
\hypersetup{
  colorlinks   = true, 
  urlcolor     = blue, 
  linkcolor    = blue, 
  citecolor   = blue 
}

\def\BibTeX{{\rm B\kern-.05em{\sc i\kern-.025em b}\kern-.08em
    T\kern-.1667em\lower.7ex\hbox{E}\kern-.125emX}}
\begin{document}

\title{USTAR: Online Multimodal Embedding for Modeling User-Guided Spatiotemporal Activity\\
\thanks{This research is financially supported by \href{https://scholarships.unimelb.edu.au/awards/graduate-research-scholarships}{Melbourne Graduate Research Scholarship} and \href{https://scholarships.unimelb.edu.au/awards/rowden-white-scholarship2}{Rowden White Scholarship}.}
}

\author{
\IEEEauthorblockN{Amila Silva, Shanika Karunasekera, Christopher Leckie, and Ling Luo}
\IEEEauthorblockA{\textit{School of Computing and Information Systems} \\
\textit{The University of Melbourne}\\
Parkville, Victoria, Australia \\
Email: \{amila.silva@student., karus@, caleckie@, ling.luo@\}unimelb.edu.au}
}

\maketitle

\begin{abstract}

Building spatiotemporal activity models for people's activities in urban spaces is important for understanding the ever-increasing complexity of urban dynamics. With the emergence of Geo-Tagged Social Media (GTSM) records, previous studies demonstrate the potential of GTSM records for spatiotemporal activity modeling. State-of-the-art methods for this task embed different modalities (location, time, and text) of GTSM records into a single embedding space. However, they ignore Non-GeoTagged Social Media (NGTSM) records, which generally account for the majority of posts (e.g., more than 95\% in Twitter), and could represent a great source of information to alleviate the sparsity of GTSM records. Furthermore, in the current spatiotemporal embedding techniques, less focus has been given to the users, who exhibit spatially motivated behaviors. To bridge this research gap, this work proposes \textsc{USTAR}, a novel online learning method for \textit{User-guided SpatioTemporal Activity Representation}, which (1) embeds locations, time, and text along with users into the same embedding space to capture their correlations; (2) uses a novel collaborative filtering approach based on two different empirically studied user behaviors to incorporate both NGTSM and GTSM records in learning; and (3) introduces a novel sampling technique to learn spatiotemporal representations in an online fashion to accommodate recent information into the embedding space, while avoiding overfitting to recent records and frequently appearing units in social media streams. Our results show that \textsc{USTAR} substantially improves the state-of-the-art for region retrieval and keyword retrieval and its potential to be applied to other downstream applications such as local event detection.

\end{abstract}


\section{Introduction}
\textbf{Motivation.} With urbanization, more than half of the today's world population (exactly 55.7\% as of 2019\footnote{\href{http://demographia.com/db-worldua.pdf}{http://demographia.com/db-worldua.pdf}}) live in urban areas. It is projected that the urbanization trend will gradually increase over the next few decades. As a result, it is not only difficult to tackle urban challenges (e.g., controlling traffic congestion, controlling environmental pollution), it is difficult for people in urban areas to find the most suitable activities and places at the right time. For instance, consider an inhabitant in a highly urbanized city like Melbourne. What is the best time to visit Mount Buller, a snowy mountain near Melbourne, for skiing? What are the popular activities or events happening around Melbourne city right now? Answering such questions is challenging due to the highly complex spatiotemporal dynamics in cities. 

Up until the early 2000s\footnote{\href{https://mktgathpu.wordpress.com/2015/06/01/the-social-media-craze-past-present-future/}{https://mktgathpu.wordpress.com/the-social-media-past-present-future}}, it was almost impossible to model these complex urban dynamics due to the lack of reliable data sources. However, with the emergence of social media and the mobile internet, people are increasingly sharing their day to day activities on social media (e.g., Twitter, Facebook) and leave a massive amount of data related to their activities. For instance, approximately 6000 tweets are posted on Twitter each second in 2019\footnote{\href{https://www.internetlivestats.com/twitter-statistics/}{https://www.internetlivestats.com/twitter-statistics/}}. Moreover, most of today's social media platforms allow users to selectively share their location information with their post, as GeoTagged Social Media (GTSM) records. Consequently, social media records have become a rich source of information to understand peoples' activities due to the ease of collecting and capturing multidimensional data (e.g., location, time, text, and user). Recent studies have demonstrated the potential of GTSM records on downstream tasks like geographical topic modeling~\cite{sizov2010geofolk,hong2012discovering,kling2014detecting}, online recommendations~\cite{zhang2017regions,zhang2017react,yang2017bridging,zhang2018spatiotemporal}, and event detection~\cite{zhang2017triovecevent,zhang2018geoburst+}. 

Our goal is to propose a unified framework to understand spatiotemporal activity dynamics in urban spaces using the power of GTSM records. Most of the earliest works on spatiotemporal modeling extend traditional topic modeling techniques to discover geographical topics from GTSM data. However, they have generally high computational complexity and make strong distributional assumptions, which dilutes the power of GTSM records to model urban dynamics. To address this, a representation learning method was firstly proposed in~\cite{zhang2017regions}, which embeds spatial, temporal, and textual units into the same latent space.
Many later studies~\cite{zhang2017regions,zhang2017react,zhang2018spatiotemporal} demonstrated that representation learning methods regularly and substantially outperform traditional geographical topic models for spatiotemporal activity modeling. However, there are several open challenges with regard to learning spatiotemporal activity representations from GTSM records.

\textbf{Research Gaps.} First, most of the previous studies on learning spatiotemporal embeddings neglect Non-GeoTagged Social Media (NGTSM) records, which is a large proportion of records compared to the GTSM records in social media platforms. For instance, less than 5\% of postings in Twitter are geotagged~\cite{chong2018exploiting,jurgens2015geolocation}. This percentage is expected to have a downward trend over the next few years as users become increasingly concerned about their privacy, which forces social media platforms to tighten their privacy agreements\footnote{\href{https://www.socialmediatoday.com/news/twitter-is-removing-the-option-to-tag-your-precise-location-in-tweets/557133/}{https://www.socialmediatoday.com/news/twitter-is-removing-tag-option}}. This will severely impact spatiotemporal modeling techniques based on just GTSM records; how data sparsity impacts the learning of state-of-the-art spatiotemporal embedding techniques is demonstrated in~\cite{zhang2018spatiotemporal}. Hence, it is important to incorporate both GTSM and NGTSM records to address the sparsity of GTSM records, despite NGTSM records being partially attributed. Nevertheless, how to jointly incorporate both GTSM and NGTSM records to learn spatiotemporal embeddings has not been well addressed.

Second, social media records generally come with the user indices, by which different users can be uniquely understood in an anonymous manner (i.e., a user index is a numerical identity given to each user as depicted in Table~\ref{tab:example1} and it does not reveal the actual identity of the user). Studies~\cite{chong2017tweet,chong2018exploiting} report that there are spatially motivated user behaviors (e.g., spatially close users produce similar textual contents and users tend to visit venues that are near to each other), which are useful to understand the dynamics of spatiotemporal units. However, such user behaviors have not been exploited to learn representations for the spatiotemporal units. 

\begin{table}[t]
\centering
\caption{A real geo-tagged tweet posted by a spectator of the Australian Open tennis tournament at the Rod Laver Arena}\label{tab:example1} 
\begin{tabular}{ll}
\hline
Location & -37.8219, 144.9785 \\
\hline
Time    & Jan 16, 2017, 15:28:07\\
\hline
UserID & 256173821 \\
\hline
Message & Andy Murray Live!!! \#melbourne \#australianopen\\ 
        & \#andymurray @ Rod Laver Arena - Australian Open\\
\hline
\end{tabular}
\vspace{-2mm}
\end{table}

Third, people's activities in the real world change over time and users usually demand recommendations that are useful at that time. For example, consider a user who queries for popular events at  Melbourne Park. Around the end of December, Christmas events happening at Melbourne Park could be a good answer to such a query. But at the end of January, the answer should reflect the Australian Open (AO) tennis tournament (i.e., one of the famous Grand Slam tennis events). 
To provide such recency-aware recommendations, the representations should be dynamically updated while giving more importance to recent information. Even though social media data are posted as a continuous stream of data, operating on them as a single batch~\cite{zhang2017regions,zhang2018spatiotemporal} does not assign more importance to the latest information. To the best of our knowledge, the 
two strategies (i.e., \textsc{ReAct-Decay} and \textsc{ReAct-Cons}) proposed in~\cite{zhang2017react} are the only works on learning recency-aware spatiotemporal embeddings. Out of the two methods, \textsc{ReAct-Decay} is the state-of-the-art, which learns spatiotemporal representations incrementally with life-decaying weights on the records to emphasize recent information. As elaborated in Section~\ref{subsec:sampling}, \textsc{ReAct-Decay} has its own limitations. Hence, sophisticated ways to accommodate online learning for spatiotemporal representation learning are yet to be explored.


\textbf{Contributions.} In this paper, we propose User-guided SpatioTemporal Activity Representation (\textsc{USTAR}), a unified framework for spatiotemporal activity modeling, which: (1) \textit{incorporates NGTSM records when learning the spatiotemporal activity model to alleviate the sparsity of GTSM records}; (2) \textit{incorporates user information for spatiotemporal activity modeling to expand the modeling potential of urban dynamics}; and (3) \textit{accommodates online learning for spatiotemporal activity modelings without overfitting to the recent information}. Our main technical contributions are as follows:

\begin{itemize}
\item \textsc{USTAR}, an embedding technique, which embeds users along with the regions, hours, and keywords into the same embedding space such that co-occurring units are close to each other in the embedding space. We show that the introduction of users into the embedding space boosts the modeling potential of urban dynamics. \textsc{USTAR} outperforms state-of-the-art methods on spatiotemporal representation learning by as much as $42.67\%$ for region retrieval and $9.62\%$ for keyword retrieval. 

\item We study two spatially motivated user behaviors in urban cities: (1) Users with similar content have similar venue histories and (2) Temporally close tweets are most likely to have similar location attributes. Based on the findings of these studies, we propose a collaborative filtering approach to infer weak geolocation labels for NGTSM records before using them jointly with GTSM records to learn \textsc{USTAR}. We show that incorporation of NGTSM records helps to alleviate the sparsity of GTSM records. 

\item We introduce a novel informative sampling technique to accommodate online learning for spatiotemporal modeling without overfitting to the most recent records in a continuous social media stream. Our sampling approach considers the informativeness of a record to the learning process in addition to its lifetime. Our results show that the proposed sampling approach outperforms existing approaches and alleviates overfitting for frequently appearing units in social media streams. 
\end{itemize}

\textbf{Paper Outline.} The rest of the paper is structured as follows. In Section~\ref{sec:related}, we discuss the previous works related to \textsc{USTAR}. Section~\ref{sec:problem} defines the problem statement. Section~\ref{sec:model} provides the technical details of \textsc{USTAR}. Section~\ref{sec:experiments} describes the experimental setup of this work. We evaluate \textsc{USTAR} in Section~\ref{sec:results} and conclude the manuscript in Section~\ref{sec:conclusion}.

\section{Related Work}\label{sec:related}
\subsection{Spatiotempral Activity Modeling.} \label{subsec:lit1}

\textbf{Probabilistic Modeling Techniques.} Existing spatiotemporal activity modeling methods can be classified into two types: topic-based~\cite{sizov2010geofolk, yin2011geographical, hong2012discovering, yin2011geographical, kling2014detecting} and embedding-based~\cite{xie2016learning, zhang2017regions,zhang2017react}. Most of the topic-based methods are extensions of standard topic models, which are based on the distribution of words. To extend such methods to the spatiotemporal domain, previous works assume that topics are not generated only based on the word distribution, but also from the distribution of locations and timestamps. For instance, Latent Dirichlet Allocation (LDA)~\cite{blei2003latent}, a famous topic modeling method, is extended in~\cite{sizov2010geofolk} assuming each latent topic has a multinomial distribution over text, and two Gaussians over latitudes and longitudes. Similarily,~\cite{yin2011geographical} extends Probabilistic Latent Semantic Analysis (PLSA)~\cite{hofmann2017probabilistic} for spatiotemporal activity modeling. However, these studies make strong assumptions on different attributes
, which might not be representative of the actual distributions in the real-world datasets.

\textbf{Representation Learning Techniques.} To mitigate the problems related to topic-based approaches, embedding-based methods have recently been proposed. \textsc{CrossMap}~\cite{zhang2017regions} is a scalable framework to learn representations for spatiotemporal activity modeling, which jointly maps different regions, hours, and keywords into the same latent space such that co-occurred units are close to each other in the embedding space. Their later work \cite{zhang2017react} proposed two different approaches (i.e., \textsc{ReAct-Decay} and \textsc{ReAct-Cons}) to learn representations in an online fashion. \textsc{ReAct-Cons} proposes a constrained based approach to accommodate recency-aware learning without overfitting. \textsc{ReAct-Decay} addresses the problem of overfitting using a time-decayed sampling approach. Generally, \textsc{ReAct-Cons} is inferior in performance to \textsc{ReAct-Decay}, which is the most relevant work for our study. 
\textsc{USTAR} differs from this work in three ways: (1) \textsc{USTAR} considers user information to guide the learning process, which is not considered by \textsc{ReAct}. Instead, \textsc{ReAct} uses partially available category labels (derived from POI checkings) to guide the learning process. However, such external labels are not usually available for all GTSM datasets; (2) \textsc{ReAct-Decay} proposes a sampling technique based on the lifetime of records to effectively incorporate recent information without overfitting. However, their approach could lead the model to overfit on relations related to popular locations and activities (elaborated in Section~\ref{subsec:sampling}). To mitigate this, \textsc{USTAR} proposes a novel informative sampling technique; and (3) \textsc{USTAR} learns from NGTSM and GTSM records to alleviate the sparsity of GTSM records, which is not addressed by \textsc{ReAct}. Their recent work \cite{zhang2018spatiotemporal} proposes \textsc{BRANCHNET} that transfers knowledge from external sources for alleviating data sparsity. However, this work still ignores the knowledge available in NGTSM records. 

\textbf{User-Guided Spatiotemporal Modeling.} Users in cities exhibit specific behaviors, which are useful to understand urban dynamics. It was shown in~\cite{chong2017tweet,chong2019fine} that similar users in urban areas tend to visit similar venues and users tend to post multiple tweets within a very short time. Following their findings, they address the prediction of tweets' geolocation using count-based approaches, which do not generalize well for unseen co-occurrences in the training phase. This is where representation learning techniques have an advantage over other methods as they capture the semantics of the units in the embedding space. In~\cite{jurgens2013s,kinsella2011m}, homophily in social networks is used to infer users' home cities, showing that users with similar home locations are relevant users. 
However, none of the aforementioned user behaviors has been exploited in existing representation learning techniques for spatiotemporal activity modeling, and all the existing user information based models are limited to specific applications (e.g., tweet geolocation prediction, users' home location prediction). To the best of our knowledge, most of the previous works based on representation learning techniques ignore user information except~\cite{zheng2012towards}, which proposes a matrix factorization based technique to learn latent representations using the co-occurrences between users, locations and activities. However, this approach is not scalable as it uses a computationally expensive matrix factorization technique. Moreover, it is difficult to learn such an approach in an online fashion. Distinguishing our work from these studies, our approach exploits user-specific behaviors to learn representations for spatiotemporal units in an online fashion, which can be applicable to a diverse set of applications.

\subsection{Online Learning}\label{subsec:lit2}
As found in previous studies, people's activities exhibit temporal dynamics:~\cite{noulas2011empirical,zhang2013importance} show that users' POI checkings have meaningful spatiotemporal patterns, which are different from region to region;~\cite{kling2012city} studies the temporal evolutions in topics by applying LDA on hourly-aggregated records as documents, and shows that there is a significant evolution in topics over time. These studies show that it is important to learn spatiotemporal activity modeling in an online fashion to capture temporal dynamics of people's activities. 

In light of that,~\cite{pozdnoukhov2011space} applies online LDA to extract topics in an online fashion, which are subsequently analyzed to understand their spatiotemporal distributions. However, this work does not provide a unified framework for spatiotemporal activity modeling. To the best of our knowledge, \textsc{ReAct} is the only previous work (compared with \textsc{USTAR} in Section~\ref{subsec:lit1}) that learns embeddings in an online fashion.  

Online learning has been previously used in other domains to model the temporal evolution of words~\cite{bamler2017dynamic}, diseases~\cite{albarqouni2016aggnet} and networks~\cite{perozzi2014deepwalk,li2017attributed}. However, \textsc{USTAR} differs from the above studies by being specific to the task of spatiotemporal activity modeling.

\section{Problem Statement}\label{sec:problem}

Let $R=\{r^1, r^2, ...., r^N, ...\}$ be a continuous stream of social media records that arrive in chronological order. There are two mutually exclusive subsets $R_{GTSM}$ and $R_{NGTSM}$  of $R$ such that $R=R_{GTSM}\cup R_{NGTSM}$. Each record $r \in R_{GTSM}$ is a quadruple $<t_r, l_r, w_r, u_r>$ and $r \in R_{NGTSM}$ is a triple $<t_r, w_r, u_r>$, where: (1) $t_r$ is the posting time of $r$; (2) $l_r$ is a two dimensional vector that represents the location of $r$; (3) $w_r$ is a bag of keywords that denotes the text message in $r$; and (4) $u_r$ is the user id of $r$.

The problem is to learn the embeddings $T$ for hours, $L$ for regions, $W$ for keywords and $U$ for users such that each element $v_x \in L \cup T \cup W \cup U$:
\begin{enumerate}
    \item is a $k$-dimensional vector ($k << |T|+|L|+|W|+|U|$), where $|T|, |L|, |W|$ and $|U|$ are the numbers of hours, regions, keywords and users in $R$, respectively;
    \item is continuously updated as new records ($R_{\Delta}$) arrive to incorporate the latest information.
\end{enumerate}

\begin{figure}[t]
    \centering
    \includegraphics[width=0.9\linewidth]{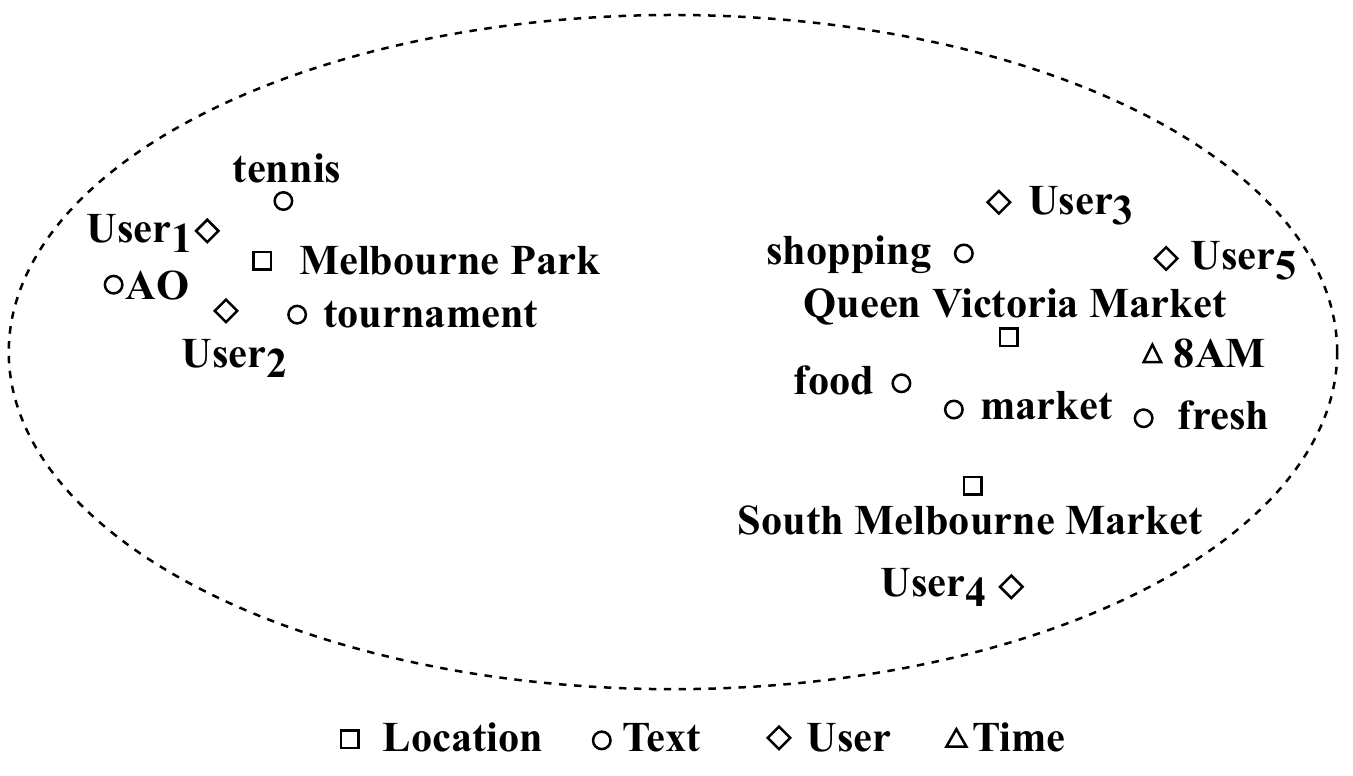}
    \caption{An illustration of the desired \textsc{USTAR} embedding space}
    \label{fig:ex}
    \vspace{-4mm}
\end{figure} 

Such a heterogeneous embedding space should be able to predict any attribute of a tweet given the other attributes (e.g., spatial query, textual query or user query). It should also be able to cluster similar values of a given attribute (e.g., locations, users or activities) together due to the similarity of their content. For example, the desired embedding space is shown in Figure~\ref{fig:ex} using a toy example. As can be seen, the co-occurring units are closely mapped in the latent space (e.g., the Melbourne Park region with the keywords ``tennis'', ``AO'', and ``tournament''); the units with the same type (i.e., location, time, text, and user)  and similar content are close to each other (e.g., Queen Victoria Market and South Melbourne Market). Following the same intuition, we can say that clustered users have similar content (e.g., User$_1$ and User$_2$).  

The heterogeneous attributes (i.e., spatial, temporal, textual and user) of $r\in R$ should be discrete and each should have a basic element to learn embeddings. In contrast, the timestamps and locations are continuous in their corresponding spaces, which should be discretized to make them feasible for embedding. Hence, following the previous works~\cite{zhang2017regions,zhang2017react}, each timestamp is converted to its corresponding hour, and geographical space is partitioned into equal-size regions.

\section{Our Approach}\label{sec:model}
\begin{figure}[t]
    \centering
    \includegraphics[width=\linewidth]{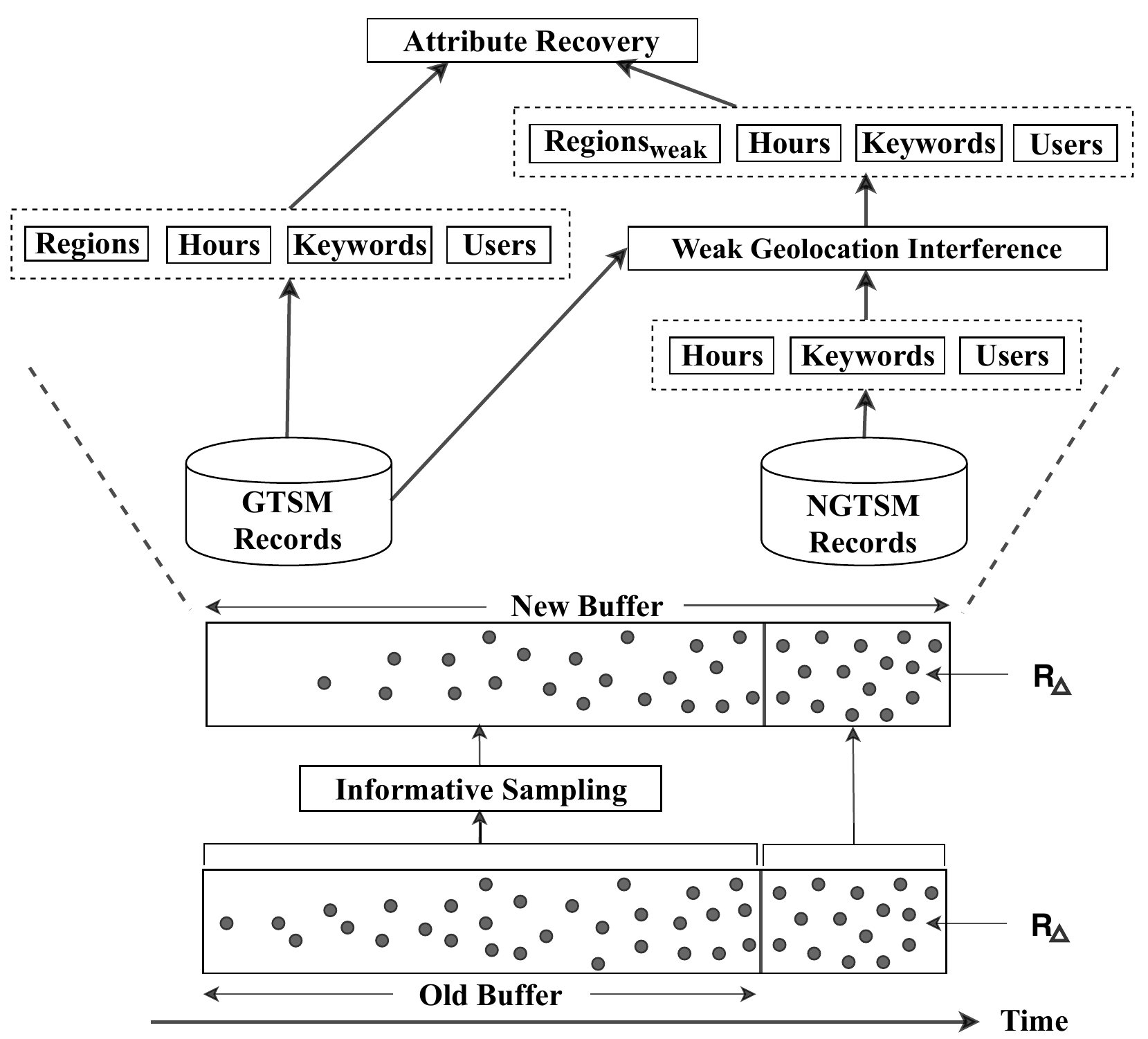}
    \caption{USTAR Framework}
    \label{fig:model}
    \vspace{-4mm}
\end{figure} 

As depicted in Figure~\ref{fig:model}, our \textsc{USTAR} model is learned in an online fashion. It gives more importance to the records from $R_{\Delta}$ to accommodate recent information to the embedding space. To avoid overfitting to $R_{\Delta}$, \textsc{USTAR} uses an informative sampling strategy to sample informative records from historical records. In subsequent sections, we verify its potential by comparing it with state-of-the-art (\textsc{ReAct-Decay})~\cite{zhang2017react}. 

For each update of the buffer $B$ (i.e., after combining $R_{\Delta}$ with the informative records from historical data), \textsc{USTAR} learns embeddings such that it recovers the attributes of $r \in B$ as much as possible. However, the NGTSM records in $B$ do not have their location attribute. If \textsc{USTAR} is trained to recover only the available attributes, the learning process of \textsc{USTAR} is biased towards recovering other attributes (i.e., time, keywords, user) except regions. We empirically observe that this distorts the embedding space of the location with the introduction of more NGTSM records to $R$. Hence, \textsc{USTAR} adopts a collaborative filtering approach to infer weak geolocation labels for NGTSM records, which is based on specific user behaviors in urbanized regions (analyzed in Section~\ref{subsec:emp_std}). This approach effectively propagates location attributes of GTSM records to NGTSM records, which are subsequently used to learn \textsc{USTAR}.

The technical details of \textsc{USTAR} (summarized in Algorithm~\ref{algo:algo1}) are elaborated in the following sections. (Note: the descriptions refer to their corresponding lines in Algorithm~\ref{algo:algo1}).

\subsection{\textsc{USTAR} Learning}\label{subsec:model_learning}
For a given record $r$, \textsc{USTAR} learns the embeddings such that the units (location, hour, user or word) of a given record $r$ can be recovered by looking at $r$'s other units\footnote{For the rest of this paper, ``units'' refers to the attribute values (e.g., Melbourne Park (location), 8AM (time), tennis (text))}. Formally, we model the likelihood for the task of recovering unit $i\in r$ given the other units $r_{-i}$ of $r$ as:
\begin{equation}
    p(i|r_{-i}) = \exp (s(i,r_{-i})/ \sum_{j\in X} \exp (s(j, r_{-i}))
\end{equation}
$X$ is the type  (could be location, time, text, or user) of $i$, and $s(i, r_{-i})$ is the similarity score between $i$ and $r_{-i}$. Following~\cite{zhang2017react}, we define the $s(i, r_{-i})$ as $s(i, r_{-i})=v_i^Th_i$ where, 

\begin{equation}
         h_i = 
\left\{\begin{matrix}
  (v_l + v_t + v_{\hat{w}}+ v_u)/4 & \quad \text{if $i$ is a keyword}\\ 
  (v_t + v_{\hat{w}}+ v_u )/3 & \quad \text{if $i$ is a region}\\ 
  (v_l + v_{\hat{w}} + v_u)/3  & \quad \text{if $i$ is an hour}\\ 
  (v_l + v_{\hat{w}} + v_t)/3  & \quad \text{if $i$ is a user}\\
\end{matrix}\right.
\end{equation}
Here $v_{\hat{w}} = \frac{1}{|W_{r_{-i}}|}\sum_{w \in W_{r_{-i}}} v_w$, $W_{r_{-i}}$ is the set of keywords in $r_{-i}$, and $v_{i}$ is the embedding of unit $i$.

Then, the final loss function for the attribute recovering task is the negative log likelihood of recovering all the attributes of the records in the current buffer $B$:
\begin{equation}\label{eq:pre_loss}
    O_{B} = - \sum_{r \in B} \sum_{i \in r} p(i|r_{-i})
\end{equation}

The objective function above is approximated using negative sampling (proposed in~\cite{mikolov2013distributed}) to efficiently optimize using stochastic gradient descent (SGD). Then for a selected record $r$ and unit $i\in r$, the loss function is:
\begin{equation}\label{eq:loss}
    L = - \log (\sigma (s(i, r_{-i}))) - \sum_{k\in K} \log (\sigma (-s(k, r_{-i})))
\end{equation}
where $\sigma(x) = \frac{1}{1 + \exp(-x)}$ and $K$ is the set of randomly selected negative units that have the type of $i$.

\begin{algorithm}[t]
 \LinesNumbered
 \SetAlgoLined
 \SetKwInOut{Input}{Input}
 \SetKwInOut{Output}{Output}
 \Input{The previous embeddings $L, T, W,$ and $U$,
       \\ The Buffer $B$,
       \\ A collection of new records $R_\Delta$.
       }
 \Output{The updated $B$ and the embeddings $L, T, W,$ and $U$.}
 \tcp{Informative Sampling}
  \For{$r$ in $B$}{
  $z_r \leftarrow Compute\_Intra\_Agreement(r)$\;
  $p_r \leftarrow \exp(-\tau z_r)$\;
  $n_{rand}\sim Bernoulli(1-p_r)$\;
  \If{$n_{rand} = 1$}{
     $B.{pop}(r)$\;
    }
   }
 $B \leftarrow B \cup R_\Delta$\;
 \For{$epoch \; from \; 1 \; to \; N$}{
 \For{$i \; from \; 1 \; to \; |R_\Delta|$}{
 $r \leftarrow \text{Sample a random record uniformly from } B$\;
 \tcp{Weak Geolocation Inference}
 \If{$r \; is \; a\; NGTSM \; record$}{
    $l_r \leftarrow Weak\_Geo\_Inference(r,U,B)$\;
 }
 \tcp{Embedding Update}
 Update L, T, W, and U to recover $r$'s attributes\;
 }
 }
 Return $B, L, T, W,$ and $U$\;
 \caption{USTAR Learning}\label{algo:algo1}
\end{algorithm}

\subsection{Informative Sampling}\label{subsec:sampling}

It is challenging to learn the aforementioned loss function using a continuous stream of data; if the records are fed to the model in their chronological order, the model could be overfitted to the most recent records~\cite{zhang2017react}. To effectively accommodate recent records, the state-of-the-art method (\textsc{ReAct-Decay}~\cite{zhang2017react}) assigns decaying weights for records based on their lifetime, such that the most recent records have higher weights. However, \textsc{ReAct-Decay} does not consider the informativeness of the record for the learning process. To exploit this, our strategy considers how much each record is learned in previous timesteps. The intuition behind our weighting scheme can be elaborated as follows.

According to our loss function in Equation~\ref{eq:pre_loss}, if the model is learned using a record $r$, $r$'s attributes should be close in the embedding space and the attributes should have a higher intra-agreement $z_r$ (denoted as $Compute\_Intra\_Agreement()$ in line 2), which is defined as:
 
 \begin{equation}
     z_r = \frac{\sum_{i, j \in r, i\neq j} \sigma(v_i^Tv_j)}{\sum_{i, j \in r, i\neq j} 1 }
 \end{equation}
 
 The informativeness of $r$ to the learning process can be defined such that it is inversely proportional to $z_r$. Since it is infeasible to store all the records and sample them based on their intra-agreement, we keep a continuously updated buffer $B$. When $R_\Delta$ arrives,  each record $r$ in $B$ is sampled for the updated buffer with a probability $p_r$ (lines 1-7):
 
 \begin{equation}
     p_r = \exp (-\tau z_r)
 \end{equation}
where $\tau > 0$ controls the decaying rate of previous samples. For NGTSM records we assign their location embeddings as identity vectors, which increases their intra-agreement and subsequently are given lesser importance by \textsc{USTAR} when sampling. This approach still gives higher importance for recent records, as the records with longer lifetimes have to go through this downsampling procedure multiple times to be kept in the buffer. 
Hence the benefits of our sampling approach are three-fold: (1) 
it still considers the recency information of a record; (2) it samples more records that are difficult to learn (records with low $z$ values); and (3) it restricts the sampling of already overfitted samples (records with high $z$ values) which subsequently reduces the possibility of the model to overfit on frequently appearing units. After downsampling $B$, $R_\Delta$ is added to the update buffer (line 9).

\subsection{Incorporating NGTSM Records}\label{subsec:coll_filt}
To infer weak geolocation labels for NGTSM records, we propose a collaborative filtering approach based on two different spatially motivated user behaviors in Twitter, which are empirically verified in the following section.

\subsubsection{\bf Empirical Analysis of User Behaviors}\label{subsec:emp_std}

Our analysis is performed using three different Twitter geotagged datasets, collected from three urbanized cities: (1)  Los Angeles (LA); (2) New York (NY); and (3) Melbourne (MB). The first two datasets (i.e., LA and NY) are the same datasets used in~\cite{zhang2017react}, the most relevant work to \textsc{USTAR}. Subsequently, we conduct all our experiments using the same three datasets. The dataset statistics are summarized in Table~\ref{tab:stat}. As can be seen, the MB dataset has fewer tweets despite having the longest collection period. Hence the information is temporally much sparser in the MB dataset compared to the other two datasets.

\begin{table}[b]
\vspace{-2mm}
\scriptsize
\centering
\caption{Dataset Statistics}\label{tab:stat} 
\begin{tabular}{|c|c|c|c|}
\hline
& LA & NY & MB\\
\hline
\hline
Tweets &1,188,405 & 1,500,000 & 263,363\\
\hline
Users    & 153,626& 176,597 & 27,056\\
\hline
Locations & 734,602& 922,845 & 30,600\\
\hline
Time Span [Month/Year] &[08/14-11/14] &[08/14-11/14] &[11/16-01/18]\\
\hline
\end{tabular}
\end{table}

\paragraph{\textbf{Spatial Homophily vs Textual Homophily}}\label{subsubsec:std1}
In this analysis, the following question is investigated: \textit{Do users with similar content history have similarity in their visit history (i.e., posting locations)?}
If this is true, the geolocation of an NGTSM record can be inferred using GTSM records, which are similar to the NGTSM record in content. This behavior was previously verified by~\cite{chong2017tweet}. Following their work, the same procedure is used to analyze the aforementioned user behavior, which is elaborated as follows:

\begin{table}[t]
\centering
\caption{Analysis on Spatial Homophily vs Textual Homophily (Section~\ref{subsubsec:std1})}\label{tab:std1} 
\begin{tabular}{|l|c|c|c|}
\hline
&  &  & p value\\
& $\overline{P_{nb}(u)}$ &$\overline{P_{nnb}(u)}$  & $H_0 :\overline{P_{nb}(u)}\le \overline{P_{nnb}(u)}$ \\
\hline
\hline
LA, $k=10$ &0.0810 & 0.0025&2.95E-211\\
\hline
LA, $k=100$ & 0.0334 & 0.0020& 2.98E-160\\
\hline
NY, $k=10$ &0.0705 & 0.0026&1.263E-165\\
\hline
NY, $k=100$ &0.0241 &0.0021 &1.16E-143 \\
\hline
MB, $k=10$ &0.1419  & 0.0453&4.69E-245\\
\hline
MB, $k=100$ & 0.1449 & 0.0447 & 4.73E-291\\
\hline
\end{tabular}
\end{table}

\begin{enumerate}
\item For each user $u$, represent $u$'s tweet content and visit history by two TFIDF vectors $I_w^u$ (considering $u$'s tweet messages as a document) and $I_l^u$ (considering $u$'s visits as a document and the locations as words) respectively.
\item Find $n$ nearest neighbors of  $u$ (denoted as $nb(u)$) based on the content using cosine similarity between $I_w^u$ and the $I_w$ of other users. Similarly, sample $n$ dissimilar users (i.e., non-neighbours) of $u$ (denoted as $nnb(u)$). 
\item Compute the visit similarity of the users with similar content as:\\ $P_{nb}(u) = \frac{1}{|nb(u)|}\sum_{u_i \in nb(u)} cos\_sim(I_l^u, I_{l}^{u_i})$. The same measure can be calculated using $nnb(u)$ to obtain $P_{nnb}(u)$.
\item For a given user set, compute the mean values of $P_{nb}(u)$ and $P_{nnb}(u)$ (denoted as $\overline{P_{nb}(u)}$ and $\overline{P_{nnb}(u)}$ respectively). Test hypothesis $H_0: \overline{P_{nb}(u)}\le \overline{P_{nnb}(u)}$ statistically (using one-tailed t test) to verify the aforementioned user behavior ($H_0$ should be rejected to verify the behavior)
\end{enumerate}

We experiment with $n = 10$ and $100$, using $5,000$ randomly selected users from each dataset, who each have at least 5 tweets. The results in Table~\ref{tab:std1} show that users' content views and venue views are positively correlated. This indicates that it is possible to exploit the content similarity between users to infer the similarity between their locations. 

\paragraph{\textbf{Spatial Homophily vs Temporal Homophily}}\label{subsubsec:std2}
In this analysis, we investigate the following question:
\textit{Do temporally close (i.e., generated within a short period) tweet messages have similar location attributes?} Such a behavior can exist due to local events and the tendency of users to move together. To explore the research question above, we conduct the following test:

\begin{enumerate}
\item Define temporally close tweets by binning tweets into a set of time bins of length $h$ based on their timestamp. 
\item Sample a random pair of tweets $(r^x, r^y)$ from a single time bin and compute the Euclidean distance between their locations $l_{r^x}$ and $l_{r^y}$ to calculate $d(r^x, r^y)$. Perform this iteratively to compute the set of $d(., .)$ for temporally close tweets (denoted as $D_{nb}$). Similarly, sample pairs of tweets from different time bins and compute the set of $d(., .)$ for temporally not close tweets (denoted as $D_{nnb}$).
\item Compute the mean values of $D_{nb}$ and $D_{nnb}$ (denoted as $\overline{D_{nb}}$ and $\overline{D_{nnb}}$ respectivly). Test hypothesis $H_0: \overline{D_{nb}}\ge \overline{D_{nnb}}$ to verify the aforementioned user behavior ($H_0$ should be rejected to verify the behavior)
 \end{enumerate}

We experiment with $h = 1$ hour and select $100,000$ random samples for each $D_{nb}$ and $D_{nnb}$. As shown in Table~\ref{tab:std2}, temporally close tweets and their location attributes are significantly correlated. This indicates that temporally close tweets are most likely generated from spatially close regions. Hence, to infer the geolocation of a tweet, other temporally close GTSM tweets could be a good search space. 

\begin{table}[t]
\centering
\caption{Analysis on Spatial Homophily vs Temporal Homophily (Section~\ref{subsubsec:std2})}\label{tab:std2} 
\begin{tabular}{|l|c|c|c|}
\hline
&  &  & p value\\
& $\overline{D_{nb}}$ &$\overline{D_{nnb}}$  & $H_0 :\overline{D_{nb}}\ge \overline{D_{nnb}}$ \\
\hline
\hline
LA & 0.1862& 0.1912 &1.21E-46\\
\hline
NY & 0.1654 & 0.1668 & 7.63E-05 \\
\hline
MB & 0.0288 & 0.0316 & 0.0\\
\hline
\end{tabular}
\end{table}

\subsubsection{\bf Weak Geolocation Inference}


Based on the findings in Section \ref{subsec:emp_std}, \textsc{USTAR} infers the geolocations of NGTSM records based on the content similarity of the users, which is discussed in detailed follows.


When looking for the weak geolocation label of an NGTSM record, it is infeasible to loop through all historical data to find the best possible weak label for the given tweet, as it requires all the previous records to be stored. Instead, we only search over the current buffer $B$, which consists of the recent collection of records $R_\Delta$ and the sampled records (refer to Section~\ref{subsec:sampling}) from the previous buffer. The intuition behind this approach is taken by the study in Section~\ref{subsubsec:std2}, which shows that temporally close tweets tend to have similar location attributes. Hence the current $B$ is a good search space to find the weak location of an NGTSM record within $R_\Delta$.

Then we use the finding in Section~\ref{subsubsec:std1} to compute the geolocation distribution of an NGTSM record using the GTSM records in $B$. The results in Section~\ref{subsubsec:std1} verifies that contentwise similar users tend to have similar visit profiles. For a given NGTSM record $r$ and a GTSM record $r'$ in $B$, we can find the contentwise similarity of the users of $r$ and $r'$ using the cosine similarity of $v_{u_r}$ and $v_{u_{r'}}$. This is because \textsc{USTAR} locates embeddings for units in a manner that preserves their co-occurrences. If two users have similar content, then they are more likely to be closely mapped in the embedding space to preserve their co-occurrences with similar content. Following that intuition, we update the geolocation distribution of $r$ as follows ($Compute\_Region\_Dist()$ in Algorithm~\ref{algo:algo2}).   

\begin{equation}
    sim_{(r', r)} = 
\left\{\begin{matrix}
  \exp(-\frac{cos\_dis(u_r, u_{r'})^2}{2c_u^2}) & cos\_dis(u_r, u_{r'}) \le c_u\\ 
  0 &  \text{otherwise}\\ 
\end{matrix}\right.
\end{equation}
\begin{equation}
    p(l_{r'}|r) = \max \{sim_{(r', r)},  p(l_{r'}|r)\}
\end{equation}
where $c_u$ controls the trade-off between the weakness of the produced labels and the number of labeled records produced by the geolocation inference approach. $cos\_dis(., .)$ denotes the cosine distance. Subsequently, the weak geolocation of $r$ is inferred by sampling from the modeled geolocation distribution $p(.|r)$, which is normalized prior to sampling via the alias sampling strategy (with $O(1)$ complexity to sample from discrete distributions) proposed in~\cite{vose1991linear}. Following that, weakly labeled NGTSM records are jointly used with GTSM records to learn \textsc{USTAR} (line 16).

\begin{algorithm}[t]
 \LinesNumbered
 \SetAlgoLined
 \SetKwInOut{Input}{Input}
 \SetKwInOut{Output}{Output}
 \Input{A NGTSM record $r$,
        \\The embeddings $U$,
        \\ The Buffer $B$.
       }
 \Output{Weak Geolocation of record $r$.}

 \tcp{Probability Distribution of Regions}
    $p(.|r) \leftarrow \emptyset$\;
  \For{$r'$ in $B$}{
  \If{$r'$ is an NGTSM record}{
  continue\;
  }
  $p(l_{r'}|r) \leftarrow Compute\_Region\_Dist(u_r, u_{r'}, p(l_{r'}|r))$
  }
  \tcp{Sampling a region from $p(.|r)$}
 $l_r \leftarrow Alias\_Sampling(p(.|r)) $\;
 Return $l_r$\;
 \caption{$Weak\_Geo\_Inference$}\label{algo:algo2}
\end{algorithm}

\subsection{Complexity Analysis}

\textbf{Space complexity. }The space complexity of 
\textsc{USTAR} consists of: (1) $\scriptstyle O(k(|L| + |T| + |W| + |U|))$ to maintain the embeddings for different modalities; and (2) $\scriptstyle O(max(|B|))$ to keep the records of the current buffer, where $\scriptstyle max(|B|) = max(|R_{\Delta}|)\frac{1}{1 -e^{-\tau}}$

\textbf{Time complexity. }The time complexity of \textsc{USTAR} consists of: (1) $\scriptstyle O(k \max(|B|))$ to perform informative sampling; (2) $\scriptstyle O(k \max(|B|)^2)$ for weak geolocation inference; and (3) $\scriptstyle O(NM^2 k \max(|B|))$ to update embeddings ($N$ and $M$ refer to the number of epochs and the maximum number of attributes of a record respectively). Our parameter study in Figure~\ref{fig:b} shows that $|B|$ in \textsc{USTAR} is much smaller compared to the buffer kept in \textsc{ReAct-Decay}. Hence, \textsc{USTAR} has a practically similar running time to \textsc{ReAct-Decay}.

\section{Experimental Setup}\label{sec:experiments}
\subsection{Datasets Construction}
All our experiments are done using the three different datasets introduced in Section~\ref{subsec:emp_std}. The datasets are preprocessed as follows. In all the tweets, we removed the stopwords, mentions, URLs, and the words that appear less than 100 times. We partition the space in each dataset into small grids with size $300m \times 300m$. The time-space is divided into $1$-hour time windows, meaning that a new set of collections are received by the model once per hour.

For comparison purposes, we use the same set of data to evaluate the performance of our approach with both GTSM and NGTSM records despite them only including geotagged tweets. To simulate NGTSM records, we keep the geolocation attribute of randomly selected $g(= \frac{|R_{GTSM}|}{|R|}\times 100\%)$ records in $R$ and remove the geolocation attribute of the rest. Our results are mainly collected with $g=50\%$ and the changes of the performance with different $g$ values are analyzed in Figure~\ref{fig:c}. 
\subsection{Baselines}
We compare \textsc{USTAR} with the following methods:
\begin{itemize}
    \item \textsc{LGTA}~\cite{yin2011geographical} is a geographical topic model that assumes each spatial region is distributed as a Gaussian distribution and each region has a multinomial distribution over words.
    \item \textsc{TF-IDF} constructs the co-occurance matrices between each pair of location, time, text and user. It calculates the tf-idf weights between each pair in the matrix by treating rows as documents and columns as words~\cite{zhang2017react}. We consider two variants of TF-IDF: 
    \begin{itemize}
        \item \textsc{TF-IDF-No-User} only considers the co-occurances between location, time, and text.
        \item \textsc{TF-IDF-User} also considers user co-occurrences to compute tf-idf similarity.
    \end{itemize}    
    \item \textsc{ReAct}~\cite{zhang2017react} is a state-of-art-method on learning recency-aware spatiotemporal activity representation. It jointly optimizes the attribute (i.e., location, time, texts, and category) recovery task and category classification task (using partially available category labels). \textsc{ReAct} proposes two different methods to accommodate recent records without overfitting them:   
    \begin{itemize}
    \item \textsc{ReAct-Cons} introduces a constraint to the objective function to avoid overfitting by preserving old embedding structure. 
    \item \textsc{ReAct-Decay} samples historical records based on their lifetime, which are subsequently added to the recent records. It learns the embeddings using the combined set of records to alleviate the overfitting to the recent information.
    \end{itemize}
\end{itemize}

\subsection{USTAR and its Variants }
Besides the aforementioned baselines, we compare \textsc{USTAR} with three variants of \textsc{USTAR}: 
\begin{itemize}
\item \textsc{USTAR} adopts an informative sampling approach (proposed in Section~\ref{subsec:sampling}) to accommodate online learning, and a collaborative filtering approach (proposed in Section~\ref{subsec:coll_filt}) to infer weak geolocation labels for NGTSM records. Both GTSM records and weakly geolocated NGTSM records are used in learning.
\item \textsc{USTAR-Decay} adopts the sampling approach in \textsc{ReAct-Decay} to accommodate online learning and the rest is similar to \textsc{USTAR}. Hence, the performance difference between \textsc{USTAR-Decay} and \textsc{USTAR} shows the importance of our informative sampling approach.
\item \textsc{USTAR-Base} uses only the GTSM records in learning. The comparison between \textsc{USTAR} and \textsc{USTAR-Base} shows the importance of NGTSM records for spatiotemporal activity modeling.
\item \textsc{USTAR-Semi} incorporates both NGTSM and GTSM records in learning, but it does not use the proposed weak geolocation inference approach in \textsc{USTAR}. Hence, \textsc{USTAR-Semi} considers only the available attributes (i.e., location, time, and text) in NGTSM records for learning. If \textsc{USTAR} is superior to \textsc{USTAR-Semi} in performance, it shows the effectiveness of our weak geolocation inference approach.
\end{itemize}


\subsection{Parameter Settings} 
All the variants of \textsc{USTAR} and \textsc{ReAct} share three common parameters (default values are shown in brackets): (1) the latent embedding dimension $k\text{ }(300)$, (2) the SGD learning rate $\eta\text{ }(0.05)$, and (3) the number of epoches $N\text{ }(50)$. We set $c=0.1$ and $\tau=1$ in \textsc{USTAR} and its variants. For the specific parameters of \textsc{ReAct}, we use the default parameters listed in~\cite{zhang2017react}. In \textsc{LGTA}, we set the number of regions to 300 and number of topics to 10 following~\cite{zhang2017react}. 

\subsection{Evaluation Tasks and Metrics}\label{sec:subsec_eval} 
Following the previous work~\cite{zhang2017react}, the performance of \textsc{USTAR} is evaluated using two different retrieval tasks: (1) region retrieval - retrieve the true region of a tweet given the other attributes of the tweet; and (2) keyword retrieval - retrieve the true keywords of a tweet given the other attributes of the tweet. For each retrieval task, we mix the ground truth with a set of $M$ negative samples to generate a candidate pool to rank. For instance, consider the region retrieval task. Given the ground truth location $l_r$ of a test instance, $l_r$ is mixed with randomly selected $M$ locations to get the candidate pool to rank. Then the size-$(M+1)$ candidate pool is ranked to get the rank of the ground truth. The average similarity of each candidate unit to the observed elements of the corresponding test query is used to produce the ranking of the candidate pool. Cosine similarity is used as the similarity measure of \textsc{USTAR} and \textsc{ReAct}. TF-IDF uses tf-idf similarity as the similarity measure. \textsc{LGTA} computes the likelihood of observing candidate attribute given the other attributes to rank candidates. The aforementioned setup is exactly similar to the one proposed in~\textsc{ReAct}

If the model is well trained to model actual spatiotemporal activities, then higher ranked units are most likely to be the ground truth. Hence, we use Mean Reciprocal Rank (MRR) as the evaluation metric to analyze the performance. For a given $Q$ number of test queries, MRR is defined as $MRR =\frac{\sum_{q = 1}^Q 1/rank_i}{|Q|}$, where $rank_i$ refers the rank of the ground truth label for the $i$-th query. This is the same evaluation metric used in~\cite{zhang2017react}.

We randomly select 20 one-hour query windows from the second half of the period for each dataset (i.e., 2014.08.01 – 2014.11.30 for LA and NY datasets and 2017.06.01 - 2018.01.31 for MB dataset), and all the tweets in the randomly selected time windows are used as test instances. 
$M$ is set to 10 for all the experiments. For each query window, we only use the tweets that arrive before the query window to train different models. Only \textsc{USTAR} and \textsc{ReAct} are trained in an online fashion and all the other baselines are trained in a batch fashion for 20 times.

\section{Results}\label{sec:results}

\begin{table}[t]
\scriptsize
\centering
\caption{The MRRs of different methods for two different retrieval tasks using GTSM records} 
\label{tab:result1} 
\begin{tabular}{lcccccc}
\hline
Method &  \multicolumn{3}{c}{Keyword Retrieval} & \multicolumn{3}{c}{Region Retrieval} \\
\hline
& LA & NY & MB & LA & NY & MB \\
\textsc{LGTA} & 0.3421 & 0.3456 & 0.3105 & 0.3682 & 0.3598 & 0.3009\\
\textsc{TF-IDF-No-User} & 0.5643 & 0.5487 & 0.5097 &0.5701 & 0.5423& 0.542 \\
\textsc{TF-IDF-User} & 0.6305 & 0.5856 & 0.5724 & 0.6517 & 0.5896 & 0.5798\\
\textsc{ReAct-Cons} & 0.6096&0.5592 & 0.5712 & 0.5989& 0.5613 & 0.5711\\
\textsc{ReAct-Decay} & 0.6171& 0.5656& 0.5769& 0.617& 0.5626 & 0.5757\\
\hline
\textsc{USTAR-Decay} & 0.6442& 0.607 & 0.6028& 0.7614& 0.775 &0.5808 \\
\textsc{USTAR} &\textbf{0.6617} & \textbf{0.6178} &\textbf{0.6324} & \textbf{0.8053}& \textbf{0.8027} &\textbf{0.6021} \\
\hline
\end{tabular}
\vspace{-2mm}
\end{table}

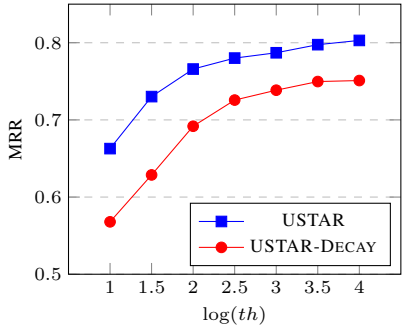
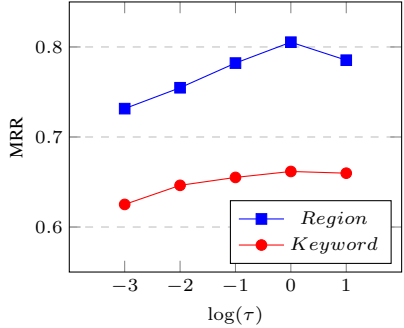
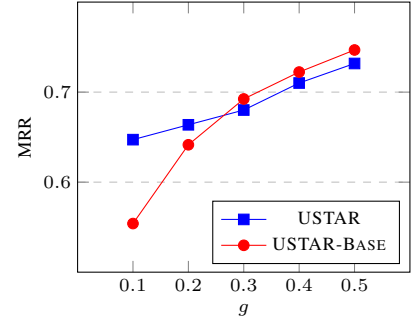
\begin{figure*}
\scriptsize
\centering
\subfloat[MRRs for retrieving regions that appear $< th$ times in the training dataset]{%
\begin{tikzpicture}
\label{fig:a}
\begin{axis}[
    xlabel={$\log(th)$},
    ylabel={MRR},
    xmin=0.5, xmax=4.5,
    ymin=0.5, ymax=0.85,
    xtick={1,1.5,2,2.5,3,3.5,4},
    ytick={0.5,0.6,0.7,0.8,0.9},
    legend pos=south east,
    ymajorgrids=true,
    grid style=dashed,
]
\addplot[
    color=blue,
    mark=square*,
    ]
    coordinates {
    (1,0.6628)(1.5,0.7302)(2,0.7659)(2.5,0.7801)(3,0.787)(3.5,0.7975)(4,0.803)
    };
    \addlegendentry{$\textsc{USTAR}$}
\addplot[
    color=red,
    mark=*,
    ]
    coordinates {
    (1,0.5679)(1.5,0.6287)(2,0.6918)(2.5,0.7257)(3,0.7385)(3.5,0.7497)(4,0.751)
    };
    \addlegendentry{$\textsc{USTAR-Decay}$}    
\end{axis}
\end{tikzpicture}
}\hspace{1em}
\subfloat[Effect of $\tau$ values]{%
\begin{tikzpicture}
\label{fig:b}
\begin{axis}[
    xlabel={$\log(\tau)$},
    ylabel={MRR},
    xmin=-4, xmax=2,
    ymin=0.55, ymax=0.85,
    xtick={ -3,-2,-1,0,1},
    ytick={0.6,0.7,0.8},
    legend pos=south east,
    ymajorgrids=true,
    grid style=dashed,
]
\addplot[
    color=blue,
    mark=square*,
    ]
    coordinates {
    (-3,0.7315)(-2,0.7547)(-1,0.7821)(0,0.8053)(1,0.7853)
    };
    \addlegendentry{$Region$} 
\addplot[
    color=red,
    mark=*,
    ]
    coordinates {
    (-3,0.6251)(-2,0.6463)(-1,0.6551)(0,0.6617)(1,0.6598)
    };
    \addlegendentry{$Keyword$}
\end{axis}
\end{tikzpicture}%
}\hspace{1em}
\subfloat[MRRs for region retrieval with different $g\%$ (percentage of GTSM records)]{%
\begin{tikzpicture}
\label{fig:c}
\begin{axis}[
    xlabel={$g$},
    ylabel={MRR},
    xmin=0, xmax=0.6,
    ymin=0.50, ymax=0.8,
    xtick={ 0.1,0.2,0.3,0.4,0.5},
    ytick={0.6,0.7},
    legend pos=south east,
    ymajorgrids=true,
    grid style=dashed,
]
\addplot[
    color=blue,
    mark=square*,
    ]
    coordinates {
    (0.1,0.6472)(0.2,0.6637)(0.3,0.6801)(0.4,0.7101)(0.5,0.7318)
    };
    \addlegendentry{$\textsc{USTAR}$} 
\addplot[
    color=red,
    mark=*,
    ]
    coordinates {
    (0.1,0.554)(0.2,0.6415)(0.3,0.6924)(0.4,0.7223)(0.5,0.7468)
    };
    \addlegendentry{$\textsc{USTAR-Base}$}
    
\end{axis}
\end{tikzpicture}
}
\vspace{-1mm}
\caption{Parameter sensitivity on LA}\label{fig:para_sens}
\vspace{-2mm}
\end{figure*}

\subsection{Quantitative Results using GTSM records}\label{subsec:geo}
Table~\ref{tab:result1} shows the results collected for the region and keywords retrieval tasks using only GTSM records. 
As shown, \textsc{USTAR} and \textsc{USTAR-Decay} show much higher MRRs than the baselines. TF-IDF is shown to be a strong baseline for the retrieval tasks. With the incorporation of user co-occurrences, it outperforms \textsc{ReAct} by on average 5.6\% for region retrieval and 1.6\% for keyword retrieval. This demonstrates the importance of user information for spatiotemporal activity modeling. 

\textsc{USTAR} outperforms \textsc{ReAct-Decay}, which is the strongest variant of \textsc{ReAct}, by as much as 42.67\% for region retrieval and 9.62\% for keyword retrieval. Since only GTSM records are used for the results in Table~\ref{tab:result1}, the significant performance improvements are caused by two different reasons.
 
First, \textsc{ReAct-Decay} does not consider the user factor, thus fails to capture spatially motivated user behaviors as empirically found in Section~\ref{subsec:emp_std}. This could be the main reason behind the significant performance boost achieved by \textsc{USTAR} especially for region retrieval.

Second, \textsc{ReAct-Decay} uses a time-decaying sampling strategy to avoid overfitting to the recent records. However, this approach could learn the model on frequently appearing units with higher importance (elaborated in Section~\ref{subsec:sampling}) and overfit to them.
Compared to that, \textsc{USTAR} proposes an approach to sample records from historical data, considering the informativeness of the records (defined based on their difficulty to learn). This argument is verified by the performance gap between \textsc{USTAR} and its variant \textsc{USTAR-Decay}, which is on average 4.34\% for region retrieval and 3.2\% for keyword retrieval. Since \textsc{USTAR-Decay} uses the same sampling strategy as \textsc{ReAct-Decay}, the proposed informative sampling approach in \textsc{USTAR} is more effective compared to the one proposed in \textsc{ReAct-Decay}. To further verify this, Figure~\ref{fig:a} shows the performance for region retrieval by \textsc{USTAR-Decay} and \textsc{USTAR}, against the number of appearances of the ground truth locations in the training set. Figure~\ref{fig:a} exhibits that \textsc{USTAR} is doing well to retrieval regions that rarely appeared in the social media records. For instance, consider the MRRs for retrieving regions that appear $<10$ times in the training dataset, \textsc{USTAR} achieves 0.6628 MRR using the LA dataset, while \textsc{USTAR-Decay} achieves 0.5679. This clearly shows that the sampling approach in \textsc{USTAR} indeed helps to learn better representations for spatial units, including the units that rarely appeared in social media.

In Figure~\ref{fig:b}, we study the effect of the $\tau$ parameter in \textsc{USTAR}. The performance first increases with $\tau$ and then becomes stable and finally reduces with further increase of $\tau$ over 10. The reason is two-fold: (1) a small $\tau$  value samples more records in history, consequently decreases the importance given to recent information; and  (2) a large $\tau$ value samples only a few samples from the history, which makes the resulting model overfit on recent records. Thus, $\tau$ controls the aforementioned trade-off to accommodate recent information effectively to learn \textsc{USTAR}.

\begin{table}[t]
\centering
\scriptsize
\caption{The MRRs for two different retrieval tasks using both GTSM and NGTSM records ($g=50\%$)}\label{tab:result2}
\begin{tabular}{lcccccc}
\hline
Method &  \multicolumn{3}{c}{Keyword Retrieval} & \multicolumn{3}{c}{Region Retrieval}\\
\hline
& LA & NY & MB & LA & NY & MB \\
\textsc{LGTA} & 0.3101 & 0.3013 &0.2984 & 0.3108 & 0.3067&0.2856\\
\textsc{TF-IDF-No-User} & 0.5521 & 0.5312& 0.4992& 0.5521& 0.5213& 0.5224\\
\textsc{TF-IDF-User} & 0.5989& 0.5642 & 0.5601& 0.6352 & 0.5622& 0.5647\\
\textsc{ReAct-Cons} & 0.5792 & 0.5403 & 0.5634 & 0.5799 & 0.5497& 0.5664\\
\textsc{ReAct-Decay} & 0.5883 & 0.5498 &  0.5661 & 0.5894 & 0.5513& 0.5672\\
\hline
\textsc{USTAR-Base} & 0.6311 & 0.5716&0.5991&\textbf{0.7485} & 0.7172&0.5923 \\ 
\textsc{USTAR-Semi} & 0.643& 0.5796& 0.6012 &0.6996 &0.7043 & 0.5925\\ 
\textsc{USTAR} &\textbf{0.6598} & \textbf{0.5805}& \textbf{0.6129}& 0.7318 & \textbf{0.7396}&\textbf{0.6177}\\ 
\hline
\end{tabular}
\vspace{-2mm}
\end{table}

\subsection{Quantitative Results using both GTSM and NGTSM records}\label{subsec:geo_nongeo}
For the results in Table~\ref{tab:result2}, 50\% of the records are randomly selected as GTSM records and the rest as NGTSM records. Thus, the performance for the retrieval tasks dropped for almost all the methods compared to the results in Table~\ref{tab:result1}. However, the ranking of the methods based on the performance is consistent over both tables. Again, \textsc{TFIDF} is shown to be a competitive baseline, which gives comparable results to \textsc{ReAct}. However, \textsc{USTAR} is the clear winner compared to the baselines, which achieves around $22\%$ and $9\%$ performance improvements over \textsc{ReAct} for region retrieval and keyword retrieval respectively.   

Comparing the variants of \textsc{USTAR}, we can see that they are inferior to \textsc{USTAR}. Another interesting observation is that \textsc{USTAR-Base} outperforms \textsc{USTAR-Semi} for region retrieval, despite that \textsc{USTAR-Base} only learns with GTSM records, which is 50\% of the records used in \textsc{USTAR-Semi}.  Although \textsc{USTAR-Semi} is built with both GTSM and NGTSM records, it only preserves the available attributes for each record. Hence, the other attributes except locations are highly emphasized in \textsc{USTAR-Semi}, which might consequently distort the location embeddings. This observation shows the importance of our weak geolocation inference algorithm. The proposed weak geolocation inference approach is the main difference between \textsc{USTAR} and \textsc{USTAR-Semi}, which accounts for nearly 4\% and 2\% performance gains for region retrieval and keyword retrieval respectively.

Surprisingly, \textsc{USTAR-Base} outperforms \textsc{USTAR} for the region retrieval task on LA. Hence, we compare \textsc{USTAR-Base} and \textsc{USTAR} for region retrieval with different $g$ values in Figure~\ref{fig:c}. It shows that \textsc{USTAR} gradually outperforms \textsc{USTAR-Base} when $g$ decreases, which signifies the importance of \textsc{USTAR} to alleviate the scarcity of GTSM records.

\subsection{Case Study}
As shown in Sections~\ref{subsec:geo} and~\ref{subsec:geo_nongeo}, \textsc{USTAR} considerably outperforms state-of-the-art methods for the region and the keyword retrieval tasks. However, \textsc{USTAR} has other potential downstream applications. We introduce an interesting case study to show a promising downstream application of \textsc{USTAR}. 

\textbf{Local Event Detection. } 
Detecting local events (i.e., unusual activity emerging in a local area) has recently been addressed due to its importance for a wide spectrum of applications (e.g., disaster control, place recommendation). A state-of-the-art method on detecting local events~\cite{zhang2017triovecevent} leverages spatiotemporal representations for local event detection, which improves the accuracy of local event detection. 

In light of this, we study the potential of \textsc{USTAR} to detect local events. To study this, we explore the temporal changes of the embedding of a given location $l$, learned via \textsc{USTAR}. We compute the temporal change $\Delta_l^t$ of $l$ at timestep $t$ as the difference of the current representation $v_l^t$ with the running mean representation of $l$. Formally, it is computed as $\Delta_l^t = euclidean\_dis(v_l^t, \frac{1}{|W|}\sum_{i=0}^W v_l^{t-k})$, where $W$ is the size of the running window and $euclidean\_dis(.,.)$ denotes the Euclidean distance function.

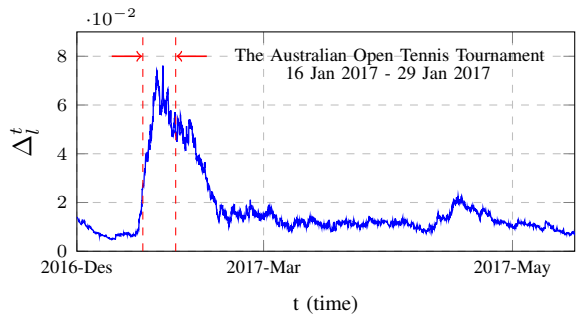
\begin{figure}[t]
\centering
\begin{tikzpicture}
\begin{axis}[
    xlabel={\footnotesize {t (time)}},
    ylabel={$\Delta_l^t$},
    xmin=0, xmax=4000,
    ymin=0, ymax=0.09,
    xtick={0,1500,3500},
    xticklabels={$\text{2016-Des}$,$\text{2017-Mar}$,$\text{2017-May}$},
    enlargelimits=false,
    legend pos=north east,
    ymajorgrids=true,
    xmajorgrids=true,
    grid style=dashed,
    width=8.2cm,
    height=4.5cm,
    tick label style={font=\scriptsize}
]
    \addplot[
    color=blue,
    each nth point={1},
    line width= 0.5pt,
    ] 
    table[x=t,y=v,col sep=comma] {data_half.csv};
    
    \addplot[dashed,red,name path=A, line width= 0.4pt
    ]coordinates {(530,0) (530,0.1)};
		 
	\addplot[dashed,red,name path=B, line width= 0.4pt
    ]coordinates {(794,0) (794,0.1)};
		 
		 
    
    
    
    \node[anchor=west] (source) at (axis cs:130,0.08){};
    \node (destination) at (axis cs:600,0.08){};
    \draw[semithick,red, ->](source)--(destination);
    
    \node[anchor=east] (source) at (axis cs:1194,0.08){};
    \node (destination) at (axis cs:724,0.08){};
    \draw[semithick,red,->](source)--(destination);
    
    \node[right] at (axis cs:1194,0.08){\scriptsize {The Australian Open Tennis Tournament}};
    \node[right] at (axis cs:1600,0.073){\scriptsize \text{16 Jan 2017 - 29 Jan 2017}};
    
\end{axis}
\end{tikzpicture}
\vspace{-1mm}
\caption{Temporal changes of the embedding of Melbourne Park}
\label{fig:case_std}
\vspace{-2mm}
\end{figure}

We select Melbourne Park as the target location for this study. We explore the temporal changes of the \textsc{USTAR} 's representation (learned using MB dataset) for Melbourne Park over time (from December 2016 to May 2017). $W$ is set to $30$ days. Figure~\ref{fig:case_std} shows the temporal changes of Melbourne Park's representation. The plot shows an unusual peak over the last fortnight of January 2017, during which the Australian Open tennis tournament was held in 2017. This indicates that \textsc{USTAR} can update the representations of locations to reflect the local events happening around it. To the best of our knowledge, none of the previous works exploit the temporal changes of spatiotemporal representation to detect local events. Thus, this could be an interesting future direction to apply \textsc{USTAR} due to its potential to accommodate recent information effectively to learn spatiotemporal modeling.

\section{Conclusion}\label{sec:conclusion}
We proposed a user-guided spatiotemporal activity model that learns representations for locations, times, texts, and users in the same latent space. Our model, \textsc{USTAR}, adapted an effective collaborative filtering based technique to incorporate both GTSM and NGTSM records to alleviate the sparsity of GTSM records. Besides, \textsc{USTAR} proposed a novel informative sampling technique to accommodate online learning for spatiotemporal activity modeling effectively.  Our results show that \textsc{USTAR} substantially outperforms the state-of-the-art and \textsc{USTAR} is useful for other downstream applications such as local event detection.

\bibliographystyle{IEEEtran}
\bibliography{references}

\end{document}